\documentclass[11pt,a4paper]{article}
\usepackage[margin=0.95in]{geometry}
\usepackage{microtype}
\usepackage[utf8]{inputenx}
\usepackage[english]{babel}
\usepackage{curve2e}
\usepackage{csquotes}
\usepackage[backend=biber, style=authoryear, citestyle=authoryear]{biblatex}
\usepackage{mathptmx}
\usepackage{helvet}
\usepackage{courier}
\usepackage{type1cm}
\usepackage{makeidx}
\usepackage{graphicx}
\usepackage{graphics}
\usepackage{multicol} 
\usepackage{amsmath}
\usepackage{booktabs}
\usepackage{float}
\usepackage[bottom]{footmisc}
\usepackage{amssymb}
\usepackage{lineno}
\usepackage{multirow}
\usepackage{wasysym}
\usepackage{tikz}
\usepackage{caption}
\usepackage{subfigure}
\usepackage{amsfonts}
\usepackage{url}
\usepackage{fancyhdr}
\usepackage{mathtools}
\usepackage{units}
\usepackage[T1]{fontenc} 
\usepackage{verbatim}
\usepackage{listings}
\usepackage{xcolor}
\usepackage{footnote}
\usepackage{enumerate}
\usepackage{ragged2e}
\usepackage{array}
\usepackage{bm}
\usepackage{calrsfs}
\usepackage{rotating}
\usepackage{tabulary}
\usepackage{blkarray}
\usepackage{bigstrut}
\usepackage{gauss}
\usepackage[ruled,vlined]{algorithm2e}
\usepackage[title]{appendix}
\usepackage{dirtytalk}
\usepackage[breaklinks=true]{hyperref}
\usepackage{amsthm}

\definecolor{lime}{HTML}{A6CE39}
\DeclareRobustCommand{\orcidicon}{%
\begin{tikzpicture}
\draw[lime, fill=lime] (0,0) 
circle [radius=0.16] 
node[white] {{\fontfamily{qag}\selectfont \tiny ID}};
\draw[white, fill=white] (-0.0625,0.095) 
circle [radius=0.007];
\end{tikzpicture}
\hspace{-2mm}
}

\foreach \x in {A, ..., Z}{%
\expandafter\xdef\csname orcid\x\endcsname{\noexpand\href{https://orcid.org/\csname orcidauthor\x\endcsname}{\noexpand\orcidicon}}
}

\newcommand{\ma}[1]{\ensuremath{\mathbf{#1}}}

\newtheorem{theorem}{Theorem}

\begin{document}
\title{A general framework for implementing distances for categorical variables}
\author{Michel van de Velden \and
Alfonso Iodice D'Enza \and
Angelos Markos \and
Carlo Cavicchia 
}



\maketitle

\begin{abstract}
The degree to which subjects differ from each other with respect to certain properties measured by a set of variables, plays an important role in many statistical methods. For example, classification, clustering, and data visualization methods all require a quantification of differences in the observed values. We can refer to the quantification of such differences, as distance. An appropriate definition of a distance depends on the nature of the data and the problem at hand. For distances between numerical variables, there exist many definitions that depend on the size of the observed differences. For categorical data, the definition of a distance is more complex, as there is no straightforward quantification of the size of the observed differences. Consequently, many proposals exist that can be used to measure differences based on categorical variables. In this paper, we introduce a general framework that allows for an efficient and transparent implementation of distances between observations on categorical variables. We show that several existing distances can be incorporated into the framework. Moreover, our framework quite naturally leads to the introduction of new distance formulations and allows for the implementation of flexible, case and data specific distance definitions. Furthermore, in a supervised classification setting, the framework can be used to construct distances that incorporate the association between the response and predictor variables and hence improve the performance of distance-based classifiers.
\end{abstract}

\section{Introduction} \label{Intro}
In many statistical methods, the quantification of dissimilarity, that is, the degree to which objects differ from each other, plays an important role. We can refer to such dissimilarity quantification as a distance. Classification methods such as K-Nearest Neighbors \parencite[KNN,][]{CH:1967}, but also clustering methods as K-means, \parencite{MacQ:1967}, partitioning around medoids \parencite[PAM,][]{KR:1990} and hierarchical linkage methods \parencite{G:1999}, and data visualization methods such as multidimensional scaling \parencite{BG:2005} and biplots \parencite{G:1971, GLLR:2011}, require a definition of distance between subjects and/or objects. The way to select a definition of distance depends on the nature of the data and problem at hand. 

Distance measures for numerical data are typically based on the magnitude of the observed differences in values \parencite[for a list of different distance measures, see, e.g.,][]{Ma:1978}. For categorical data, however, the situation is more complex, as we do not directly observe, and hence cannot directly quantify, sizes of differences. We can only directly establish whether there is a difference or not. 

For distance calculations in multivariate contexts, two cases can be distinguished. First, the distances are calculated for each variable independently and then added. Second, the association between the variables is taken into account when calculating the distances. For numerical variables, several well-known distances, for example, Euclidean or Manhattan distances, implicitly assume independence between the variables. Obviously, in such \say{independent} cases, the measurement scales must be commensurable. For categorical variables, there are also several measures that take the sum of dissimilarities per variable when considering a multivariate distance. For example, in simple matching, distance between two observations is defined as the number of times that the categories of corresponding variables do not match. 

For numerical variables, the association between variables can be accounted for using the Mahalanobis distance, where (sample) covariances are used to weigh observed differences to account for correlation between the variables. For categorical data, so-called association-based distances exist. In such distances, the association between categorical variables is used to quantify differences between observations. The question of how to account for associations in a categorical setting is not trivial. 
Several relatively recent new proposals for distances between categorical variables are indeed association-based distances \parencite[see, e.g.,][]{LH:2005,AD:2007,JCL:2014,ROBNLH:2015}.

The complexity of defining a distance for categorical variables, and recent interest in this topic, is illustrated by a wide range of articles that review existing \parencite[e.g.,][]{BCK:2008, ACN:2019} or introduce (new) distances \parencite[e.g.,][]{LH:2005, AD:2007,JCL:2014,ROBNLH:2015,SR:2019,BL:2022}. In this paper, we propose a general framework for implementing categorical distances. Our framework can be used to incorporate existing categorical variable distances, but it also allows researchers to define and implement new or customized distances. 

By reformulating existing distances in our framework, it becomes possible to assess the differences and similarities between them. Currently, such comparisons are not trivial due to the wide variety of notation and research fields (and hence objectives) in which methods have been proposed. In addition, our framework makes it possible to construct and define new and highly customizable distances. For example, in a supervised classification context, the framework can be used to define a distance that takes into account association with the classes of the response variable. 

As our framework is not method- or application-specific, it can be used to calculate distance matrices for any method or application requiring distance calculations. For example, multidimensional scaling, cluster analysis, or, in a supervised context, K-nearest neighbors. We show that distance calculations using the framework are fast, efficient, and transparent and can be a significant improvement over existing implementations. In particular, we show that for the distance for categorical variables proposed in \textcite{AD:2007}, our implementation is much faster than existing implementations.

An important issue with regard to the definition of distance measures is the validation of the measures. That is, how does one know that the chosen measure is appropriate? Although the new framework does not provide an answer to this question, having one general formulation simplifies both theoretical and empirical comparisons between different choices. 

We illustrate our method by applying distance-based data analysis methods to several well-known categorical data sets using a selection of known and new, association-based, categorical distances. We implemented functions to perform all categorical distance calculations using our general framework in the R package \texttt{catdist}, which is available on GitHub\footnote{\url{https://github.com/alfonsoIodiceDE/catdist_package}} and is soon to be released on CRAN. 

This paper is organized as follows. After introducing some notation in Section \ref{SectFrame}, we describe our general framework in Section \ref{SectCatDistances}. In Section \ref{Sect:IndependentDistances}, we introduce several common distance measures for categorical variables and show how they can be incorporated. Categorical distances based on co-occurrences are introduced in Section \ref{SectAssociationBasedDist}, with particular attention to a distance measure proposed by \textcite{AD:2007}. In Section \ref{SectSupervised}, we show how supervised distances can be constructed and implemented using our framework. Tuning of distance definitions is described in Section \ref{Sect:Tuning}, after which we illustrate our methodology using several data sets in Section \ref{SectApplications}. Section \ref{SectConclusion} concludes the paper. 

\section{Notation} \label{SectFrame}
Suppose that we have $n$ observations on $Q$ categorical variables and let the number of categories for the $j \in 1 \dots Q$-th variable be $q_{j}$. We can then code the categorical data by using indicator matrices. That is, for each categorical variable $j \in 1 \dots Q$, we create an $n \times q_j$ binary matrix $\ma{Z}_{j}$, where the $n$ rows correspond to observations and the $q_j$ columns to categories. The observed category is indicated by a one, and all other categories are assigned zeros. Furthermore, for each observation of a categorical variable, exactly one category is observed, and we only include categories that have been observed at least once in the data set. Hence, each column of $\ma{Z}_{j}$ contains at least one element equal to one and $\ma{Z}_{j}\ma{1}_{q_{j}}=\ma{1}_{n}$, where, generically, $\ma{1}_{i}$ denotes an $i$ by $1$ vector of ones. That is, the sum over the columns is $1$. 

Using these indicator matrices, we can code data on $Q$ categorical variables in a so-called super-indicator matrix by collecting all indicator matrices next to each other. That is,
\begin{align*}
\ma{Z}=\left( \begin{array}{ccc} \ma{Z}_1 & \dots & \ma{Z}_Q 
\end{array} \right).
\end{align*}

Furthermore, define
\begin{align}\label{P}
\ma{P}=\frac{1}{n}\ma{Z}^{\prime}\ma{Z},
\end{align}
and
\begin{align}\label{Pd}
\ma{P}_d = \frac{1}{n} \left(\ma{Z}^{\prime}\ma{Z}\right) \odot \ma{I}_{Q^*},
\end{align}
where $\odot$ indicates the Hadamard product, that is, element-wise multiplication, and $Q^{*}=\sum_{j=1}^{Q} q_j$. Note that $\ma{P}_d$ is a diagonal matrix with as its diagonal elements the observed relative frequencies (within each variable) for the categories. Moreover, let 
\begin{align} \label{p}
\ma{p}=\ma{P}_d \ma{1}_{Q^*}
\end{align}
denote the vector of observed relative frequencies, and 
\begin{align} \label{pinv}
\ma{p}^-=\ma{P}_{d}^{-1} \ma{1}_{Q^*}
\end{align}
is the vector of inverse observed relative frequencies. 

Note that the $ij$-th off-diagonal block of $\ma{P}$ gives the relative frequencies of co-occurrences for the categories of variables $i$ and $j$. They can be seen as (empirical) joint probability distributions for variables $i$ and $j$. For the calculation of association-based distances in Section \ref{SectAssociationBasedDist}, we also define
\begin{align}\label{R}
\ma{R} = \ma{P}_{d}^{-1}\left( \ma{P} - \ma{P}^{}_{d} \right).
\end{align}
The rows of the $ij$-th off-diagonal block of $\ma{R}$ give, for the categories of the $i$-th variable, the distributions over the categories of the $j$-th variable. These can be interpreted as (empirical) conditional distributions. 

\section{Categorical distance calculations based on category dissimilarities} \label{SectCatDistances}
For a categorical variable, it is not obvious how to quantify differences between different categories. For example, suppose that we observe three individuals, one from the Netherlands, one from Italy, and one from Greece. Geographically, and perhaps also culturally, Italy and Greece are more similar than the Netherlands. How to take such differences into account is, however, not trivial. In our framework, we do so by defining {\em category dissimilarities}. 

A matrix $\bm{\Delta}_{j}$ is the category dissimilarity matrix for variable $j$. The elements of this matrix, $\delta_{ab}$, where $a$ and $b$ indicate two categories of variable $j$, quantify the dissimilarities between the categories $a$ and $b$ of the $j$-th variable. 
We can impose conditions on the dissimilarity matrix that are consistent with typical distance definitions. That is, 1) the dissimilarity of a category from itself is zero ($\delta_{aa} = 0$, for all categories). 2) Dissimilarities are symmetric ($\delta_{ab} = \delta_{ba}$, for all pairs of categories). 3) Dissimilarities satisfy the triangle inequality. That is, if $a, b$ and $c$ denote different categories for a variable $j$, then, for all categories $a, b$ and $c$,
\begin{align*}
\delta_{ac} \leq \delta_{ab} + \delta_{bc}. 
\end{align*}
If all three of these conditions are satisfied, the dissimilarities can be considered as metric distances between categories. If they are non-negative and satisfy only the first two conditions, they can be interpreted as non-metric distances between categories. However, we refer to them as category dissimilarities and reserve the term \say{distance} for the distances between observations.

If we have $Q$ categorical variables, each with a category dissimilarity matrix $\bm{\Delta}_{j}$, we can construct a $Q^{*} \times Q^{*}$, block diagonal matrix $\bm{\Delta}$, with separate category dissimilarity matrices as diagonal blocks. 

The category dissimilarity matrices can be used to calculate a between observations distance matrix as follows. First, consider the $n$ by $q_j$ indicator matrix $\ma{Z}_j$ corresponding to the $j$-th categorical variable. Furthermore, we have the corresponding category dissimilarity matrix $\bm{\Delta}_j$. We can formulate the following theorems:

\begin{theorem}\label{T2}
The distances between the observations for the categorical variable $j$ are
\begin{align*}
\ma{D}^{}_{j}=\ma{Z}^{}_{j} \bm{\Delta}^{}_{j}\ma{Z}_{j}^{\prime}.
\end{align*}
\end{theorem}

\renewcommand\qedsymbol{$\blacksquare$}
\begin{proof}
The matrix multiplication of the row $i$ of $\ma{Z}_{j}$ with $\bm{\Delta}_{j}$ selects the row of $\bm{\Delta}_{j}$ corresponding to the category chosen by the individual $i$. Similarly, matrix multiplication of this row by the $i^{\prime}$-th column of $\ma{Z}_{j}^{\prime}$ (i.e., the $i^{\prime}$-th observation) selects the element corresponding to the category chosen by the individual $i^{\prime}$. Hence, the $(i,i^{\prime})$-th element of $\ma{D}_{j}$ is the dissimilarity between the categories chosen by individuals $i$ and $i^{\prime}$. 
\end{proof}

\begin{theorem}
If we define the distance between observations on $Q$ categorical variables as the sum of $Q$ dissimilarities for each categorical variable, the $n \times n$ distance matrix can be calculated as
\begin{equation}\label{ZDZ}
\ma{D}= \ma{Z}\bm{\Delta}\ma{Z}^{\prime}.
\end{equation}
\end{theorem}
\begin{proof}
\begin{align*}
\ma{D}& = \ma{Z}\bm{\Delta}\ma{Z}^{\prime} \\
& = \left(
\begin{array}{ccc} \ma{Z}_{1} & \dots & \ma{Z}_Q \end{array} \right)\left(\begin{array}{ccc}
\bm{\Delta}_1  & & \\ 
& \ddots &\\
& & \bm{\Delta}_Q \end{array} \right) \left( \begin{array}{c}
\ma{Z}_{1}^{\prime}\\
\vdots \\
\ma{Z}_{Q}^{\prime}
\end{array} \right) \\ 
&= \sum_{j=1}^{Q} {\ma{Z}^{}_{j}\bm{\Delta}^{}_{j}\ma{Z}_{j}^{\prime}}\\
&= \sum_{j=1}^{Q}{\ma{D}_{j}}
\end{align*}
\end{proof}

From (\ref{ZDZ}), it follows that distances between observations of categorical variables depend on the choices of the category dissimilarity matrices $\bm{\Delta}_{j}$. This allows for great flexibility in defining a suitable distance measure for a set of categorical variables. In the next section, we briefly review some choices for $\bm{\Delta}$, and we show how they relate to existing distances. 


Note that associations between categorical variables are not explicitly incorporated in this formulation. That is, the differences between the categories observed for one variable are not related to the differences in the categories observed for other variables. There are, however, ways to account for such observations. For example, rather than creating an indicator matrix for each categorical variable, one could construct an indicator matrix for all possible combinations (or subsets thereof) of observations. That is, one can create, for each (or a subset of) combination of categories one indicator matrix. The number of columns of such a matrix is therefore $\prod_{j=1}^{Q}{q_j}$ and only one category dissimilarity matrix is needed where each category is a combination of the categories for all $Q$ variables. However, with several categorical variables, the total number of combinations and hence the number of categories of the final indicator matrix quickly becomes large. Furthermore, finding an appropriate category dissimilarity matrix for the combinations is not a trivial task. 

An alternative way to account for associations between the categorical variables is to use them in the construction of the category dissimilarity matrices. That is, by defining the dissimilarities between the categories of a variable in $\bm{\Delta}_j$, based on the associations with other variables. In Section \ref{SectAssociationBasedDist}, we give examples of such category dissimilarity measures. 

\subsection{Distances between sets} \label{dist_sets}
Suppose that we have two separate sets of observations on the same $Q$ categorical variables. Data for these two sets can be collected in the $n_1 \times Q^*$ and $n_2 \times Q^*$ super indicator matrices $\ma{Z}^{(1)}$ and $\ma{Z}^{(2)}$. Then, for a known category dissimilarity matrix $\bm{\Delta}$, it is easily verified that the distances between the observations for the two sets can be calculated as
\begin{align}\label{BetweenSetDist}
\ma{D}^{(12)} = \ma{Z}^{(1)} \bm{\Delta}\ma{Z}^{(2)\prime}.
\end{align}
Note that the matrix $\ma{D}^{(12)}$ is of order $n_1 \times n_2$. 

The calculation of distances between sets can be useful when considering distance-based classification problems. In KNN, for example, distances between \say{new} (unlabeled) observations and observations in a labeled data set are required. The KNN predictions are based on (usually by majority vote) the labels of the $K$ nearest neighbors in the training set. Similarly, in partitioning around medoids, a popular clustering algorithm similar to K-means, where instead of considering within-cluster variation around the mean, variation around an actual observation, the medoid, is considered. If the medoids are collected in $\ma{Z}^{(1)}$ and \say{new} data points in $\ma{Z}^{(2)}$, we can assign the new points to existing clusters considering the distances $\ma{D}^{(12)}$ and selecting the smallest distances.

\section{Independent category dissimilarity matrices} \label{Sect:IndependentDistances}
We first consider several definitions of dissimilarity for categorical data that do not take into account the association between variables. Hence, in a multivariate context, distances are calculated as the sum of distances per variable, and for each variable, the category dissimilarities are independent of the observations on other variables. However, category dissimilarities may depend on the observed frequencies for a variable. 

We do not aim to be complete with respect to the different definitions of dissimilarity. Instead, we select definitions from \textcite{SR:2019} \parencite[which contain several definitions also reviewed in][]{BCK:2008, ACN:2019}, and transform them into dissimilarities. We show how these dissimilarities can be incorporated into our framework by defining the appropriate category dissimilarity matrices $\bm{\Delta}$. For a more detailed description, as well as study on the relative performances of these dissimilarity definitions in a cluster analysis setting, see \textcite{SR:2019}. 

\subsection{Overlap or simple matching} \label{Sect:SM}
The idea of simple matching (SM) is that the distance between observations is $1$ if the categories do not match and $0$ if they do. Consequently, all different between category dissimilarities are $1$. For the $j$-th categorical variable with $q_j$ categories, we define
\begin{align*}
\bm{\Delta}^{}_{{M}_j} = \ma{1}^{}_{q_j} \ma{1}_{q_j}^{\prime} - \ma{I}^{}_{q_j}.
\end{align*}
That is, the distance between each category is exactly $1$. If we have $Q$ categorical variables and want to calculate a simple matching for all variables, we simply collect all $\bm{\Delta}_{{M}_j}$ in a block diagonal matrix $\bm{\Delta}_M$. 
\begin{align*}
\bm{\Delta}_{M} = \ma{K}_b - \ma{I}_{Q^*},
\end{align*}
where $\ma{K}_b$ is a $Q^{*} \times Q^{*}$ block diagonal matrix with, for $j=1 \dots Q$, $q_j \times q_j$ matrices ($\ma{K}_{b_j}$) of ones as its diagonal blocks.

\subsection{Eskin} \label{Sec:Eskin}
For Eskin distance \parencite{EAPPS:2002}, category dissimilarities depend on the number of categories. Dissimilarities for variables with more categories are smaller than dissimilarities for variables with fewer categories. In particular, the dissimilarity between different categories for a variable with $q_j$ categories is $2/q^2_j$. Therefore, for the $j$-th categorical variable with $q_j$ categories, the category dissimilarity matrix is defined as
\begin{align*}
\bm{\Delta}^{}_{E_j} = \left(2/q_j^2 \right)\left( \ma{1}^{}_{q_j}\ma{1}_{q_j}^{\prime} - \ma{I}^{}_{q_j}\right). 
\end{align*}
Collecting all $Q$ dissimilarity matrices in a block diagonal matrix produces the Eskin category dissimilarity matrix $\bm{\Delta}_{E}$. If all variables have the same number of categories, Eskin merely re-scales the simple matching dissimilarity.

\subsection{Lin}  \label{Sec:Lin}
\textcite{L:1988} proposed an information-theoretic measure
that gives more weight to matches on frequent values and lower weight to mismatches on infrequent values. We implement Lin's proposal as follows: define $\ma{P}_r=\ma{p} \ma{1}^{\prime}_{Q^*}$ and $\ma{P}_c=\ma{1}_{Q^*}\ma{p}^{\prime}$, where $\ma{p}$ is as defined in \ref{p}. Furthermore, let $\widetilde{\ma{P}}= \ma{P}_r + \ma{P}_c - \ma{P}_d$. Then, the category dissimilarity matrix can be defined as
\begin{align*}
\bm{\Delta}_{Lin} = \left[\log(\ma{P}_r)+\log(\ma{P}_c)-2\log(\widetilde{\ma{P}}) \right] \oslash 2\log(\ma{P}_r + \ma{P}_c),
\end{align*}
where $\oslash$ indicates the Hadamard division (i.e., element-wise) and $\log\left(\ma{\cdot}\right)$ takes the logarithms of the elements of the parenthesized object and collects them in an object of the same size. 

Note that Lin's dissimilarity for a category with itself is zero. Furthermore, in our implementation, for each variable, the dissimilarity between different categories, say categories $a$ and $b$, is 
\begin{align*}
\left[\log(p_a) + \log(p_b) - 2\log(p_a + p_b)\right]/2\log(p_a + p_b),
\end{align*} 
where $p_a$ and $p_b$ are, respectively, the relative frequencies of categories $a$ and $b$. 

\subsection{Inverse occurrence frequency} \label{Sec:IOF}
For inverse occurrence frequency \parencite[IOF,][]{BCK:2008}, a higher dissimilarity is assigned when categories are more frequently observed. In particular, the category dissimilarity matrix is defined as
\begin{align*}
\bm{\Delta}_{IOF} = \left[\log\left(n\ma{p}\right)\right]\left[\log\left(n\ma{p}\right)\right]^{\prime}- \left[\log\left(n\ma{p}\right)\right]\left[\log\left(n\ma{p}\right)\right]^{\prime}\odot \ma{I}_{Q^*}.
\end{align*}

It is worth observing that, for each variable, IOF dissimilarity for a category with itself is zero, and the dissimilarity between two different categories, say $a$ and $b$, corresponds to
\begin{align*}
\log(n p_a) \log(n p_b).
\end{align*} 

The IOF measure is related to the concept of inverse document frequency (TF-IDF) from information retrieval, where it is used to account for document relevance for a given term \parencite{S:1972}. In other words, since a rare term contributes more information than a more frequent term, the IOF measure accounts for how rare the term is, and a lower IOF dissimilarity corresponds to a rarer term. Log frequency is used to reduce the impact of terms of very high frequencies.

\subsection{Occurrence frequency} \label{Sec:OF}
For occurrence frequency (OF) dissimilarity, dissimilarities are higher if the categories are observed less frequently. The category dissimilarity matrix is defined as
\begin{align*}
\bm{\Delta}_{OF} = \left[\log\left(\ma{p}^{-}\right)\right]\left[\log\left(\ma{p}^{-}\right)\right]^{\prime}- \left[\log\left(\ma{p}^{-}\right)\right]\left[\log\left(\ma{p}^{-}\right)\right]^{\prime}\odot \ma{I}_{Q^*}.
\end{align*}

Therefore, OF dissimilarity for a category with itself is zero, and the dissimilarity between two different categories $a$ and $b$ is 
\begin{align*}
\log(p_a) \log(p_b).
\end{align*} 

\subsection{Goodall dissimilarities} \label{Sect:Goodall}
In \textcite{BCK:2008}, four variations of Goodall's similarity are considered. These are based on Goodall's original proposal \parencite{G:1966}. After transforming similarities into dissimilarities, where dissimilarity is $1 -$ similarity, the four measures have in common that dissimilarities between different categories are, as is the case with simple matching, always equal to one. However, the dissimilarity of a category with respect to the same category depends on the observed proportions of the categories. Below we provide the category dissimilarity matrices for Goodall 3 and Goodall 4. For Goodall 1 and 2, we can also construct such matrices. However, these definitions require conditional sums of proportions. In particular, for Goodall 1, the dissimilarity for category $a$ with itself, is defined as the sum of squared observed proportions that are smaller or equal to the observed proportion of category $a$. For Goodall 2, it is the sum of squared observed proportions that are larger or equal to the observed proportion of category $a$. 

The Goodall 3 and 4 measures do not require the calculation of a (conditional) sum and have the squared proportion and one minus the squared proportion of a category, respectively, on the diagonal blocks of $\bm{\Delta}$. That is, 
\begin{align*}
\bm{\Delta}^{}_{G_3} = \ma{K}^{}_b - \ma{I}^{}_{Q^*} + \ma{P}^{2}_d,
\end{align*}
and
\begin{align*}
\bm{\Delta}^{}_{G_4} = \ma{K}^{}_b - \ma{P}^{2}_d.
\end{align*}
In these definitions, dissimilarity of a category with the same category is not zero. Consequently, the resulting \say{distances} do not satisfy the typical requirements of a distance. Note that for the Goodall 1 and 3 measures, a higher dissimilarity is assigned when the matching categories are frequent, whereas for the Goodall 2 and 4 measures a higher dissimilarity is assigned when the matching categories are infrequent.

\subsection{Variable Entropy and Variable Mutability dissimilarities} \label{Sec:VEVM}
\textcite{SR:2019} proposed two variability-based dissimilarity measures that are related to Goodall 1 and 2, respectively. These dissimilarities are equal to one if the categories do not match, while, if they do match, the Variable Entropy (VE) measure uses the entropy and the Variable Mutability (VM) measure uses the Gini coefficient to quantify \say{dissimilarity}. In particular, for the $j$-th categorical variable with $q_j$ categories, the category dissimilarity matrices are defined as 
\begin{align*}
\bm{\Delta}_{VE_j} = \ma{K}_{b_j} + \left(\frac{1}{\log q_j} \sum_{l=1}^{q_j}{p_{l}\log p_{l}}\right) \ma{I}_{q_j}
\end{align*}
and 
\begin{align*}
\bm{\Delta}_{VM_j} = \ma{K}_{b_j} - \left[\frac{q_j}{q_j - 1} \left(1- \sum_{l=1}^{q_j}{p_{l}^2} \right)\right] \ma{I}_{q_j},
\end{align*}
respectively. Collecting all $Q$ dissimilarity matrices in a block diagonal matrix returns the VE and VM category dissimilarity matrices $\bm{\Delta}_{VE}$ and $\bm{\Delta}_{VM}$.

\subsection{Ordered categories} \label{Sect:order}
If categories are ordered, the order can be reflected in the dissimilarities. A simple choice would be to define the dissimilarities as the difference in category numbers. That is, the dissimilarity between categories $a$ and $b$ is simply $b-a$. If the data are rank order data or rating (e.g., Likert) scale data, this definition would imply treating the data as interval data. However, implementation of alternative, custom, definitions of ordered between-category distances is also straightforward. For example, if the categories correspond to bins on a numerical scale (e.g., age or income groups), differences between the midpoints of the bins can be used to define dissimilarities that better reflect the underlying values. More generally, let $\bm{\Delta}^{i}_{o}$ denote the $i$-th diagonal block of $\bm{\Delta}^{*}_{o}$, and $\delta^{i}_{ab}$ its $ab$-th element. Then, dissimilarities between ordered categories can be imposed by letting $\delta^{i}_{ab} \leq \delta^{i}_{ab^*},$ for $a,b \in 1 \dots i_{q_i}$, $b^* \neq b$ and $b^* > a$. Note that this definition does not guarantee that the triangle inequality holds. That is, without additional constraints, it may be the case that the direct distances between two categories are larger than the indirect distances between those categories.

\section{Association-based category dissimilarity matrices} \label{SectAssociationBasedDist}
Several authors \parencite[e.g.,][]{AD:2007, JCL:2014, LH:2005, ROBNLH:2015} have proposed distance measures for categorical variables that take into account the association between categorical variables. The general idea is that, similar to the case of the Mahalanobis distance for numerical variables, differences that are in line with the association between variables are less informative (i.e., should correspond to smaller dissimilarity values) than differences that are not in line with the general association. How to exactly implement this idea depends on the calculation of the association between categorical variables, and how to incorporate this association in the category dissimilarities. 

Here, we present a general form to calculate and collect association-based dissimilarities that can be directly implemented in our general framework in Section \ref{SectCatDistances}. We then present some specific variants and link them to recent proposals. 

\subsection{A general form for association-based dissimilarities} \label{Sect:AssociationBasedDistGenform}
Fundamental in the calculation of association-based distances is the matrix of proportions of co-occurrences $\ma{P}$ and the corresponding profile matrix $\ma{R}$ as defined in Equations (\ref{P}) and (\ref{R}). In particular, recall that the off-diagonal blocks of $\ma{P}$ and $\ma{R}$ can be interpreted as (empirical) joint and conditional probability distributions, respectively. By considering different ways to quantify the dissimilarities between the conditional distributions (i.e., the rows of the off-diagonal blocks of $\ma{R}$) we can construct different category dissimilarity matrices $\bm{\Delta}$ that, by applying Equation (\ref{ZDZ}) can be used to obtain the between-observation distances. 

Let $\ma{R}^{ij}$ denote the $ij$-th off-diagonal block of $\ma{R}$, and let $\ma{r}_{a}^{ij}$ denote its $a$-th row. Note that the elements of each row of $\ma{R}^{ij}$ add up to $1$. Hence, these elements can be seen as (empirical) conditional probabilities. We define the dissimilarities between categories for all pairs of categories of variable $i$ (for $i=1 \dots Q$) based on the association with variable $j$ (with $j \neq i$), as
\begin{align}\label{Phi_ij}
\delta^{ij}(a,b)= \Phi^{ij}\left(\ma{r}_{a}^{ij},\ma{r}_{b}^{ij}\right),
\end{align} 
where, generically, $a$ and $b$ indicate categories of variable $i$ and $\Phi^{ij}$ is the dissimilarity function that quantifies the differences between profiles based on the association between variables $i$ and $j$. The overall between category dissimilarities for all pairs of categories of variable $i$ can be defined as
\begin{align} \label{genformula}
\delta^{i}(a,b)=\sum_{j \neq i}^{Q}{w_{ij}\delta^{ij}(a,b)}= \sum_{j \neq i}^{Q}{w_{ij}\Phi^{ij}\left(\ma{r}_{a}^{ij},\ma{r}_{b}^{ij}\right)}.
\end{align}
The weights $w_{ij}$ in Equation (\ref{genformula}) allow flexibility with respect to the importance of different variables in the calculation of association-based category dissimilarities, as defined by $\Phi^{ij}$. By collecting, for each variable $i$, the elements $\delta^{i}(a,b)$ in a category dissimilarity matrix $\bm{\Delta}^{i}_{\Phi}$, and organizing them on the diagonal of a block diagonal matrix, we obtain a dissimilarity matrix $\bm{\Delta}_{\Phi}$, which can be used to calculate the distances between the observations using Equation (\ref{ZDZ}). 

Equation \ref{genformula} provides a very general way to define category dissimilarities using pair-specific weights and dissimilarity functions.

For the association-based dissimilarity functions $\Phi^{ij}$, any function that quantifies the difference between two distributions can be used. A brief overview of 46 different functions and their implementation in the R package \texttt{philentropy} is described in \textcite{D:2018}, and a more comprehensive overview of those functions, dividing them into different types and classes, can be found in \textcite{C:2007}. 

Concerning the choice of weights $w_{ij}$, we distinguish two options. In the first, all weights are equal. Usually $1$ or $1/(Q-1)$, so that either sums or averages are obtained. Alternatively, different weights can be used for different pairs. These weights can either be selected using expert knowledge (e.g., based on the experience and preferences of the researcher) or by using a data-driven approach. For example, one could set certain weights to zero and others to some constant based on some predetermined data dependent criterion (e.g., a measure of association like Cram\'er's $V$). Pairs with non-zero weights can then be referred to as \say{context} variables. Approaches using such context-based dissimilarities are described in \textcite{IPM:2009, JCL:2014, ROBNLH:2015}.

If an objective measure of overall fit of a solution is available, one could consider $w_{ij}$ and $\Phi^{ij}$ as tuning parameters, and search combinations of these parameters to make a choice.
In Section \ref{Sect:Tuning} we shall further explore the tuning of $w_{ij}$ and $\Phi^{ij}$. 

In the next subsections, we describe some specific choices of dissimilarity functions. In particular, we provide definitions for category dissimilarities between categories $a$ and $b$ of variable $i$, based on the association between variables $i$ and $j$. That is, we present specific choices of $\Phi^{ij}$ in Equation (\ref{Phi_ij}). Inserting these definitions into Equation (\ref{genformula}) results in a dissimilarity matrix $\bm{\Delta}_{\Phi}$ that can be used to calculate the between-observation distances. For ease of notation, we now drop the superscripts $ij$.



\subsection{Total variation distance between profiles} \label{Sect:TVD}
The total variation distance (TVD) between two discrete probability distributions 
can be defined as $1/2$ times the $L_1$ norm between the distributions. We can implement this in our framework by defining the category dissimilarity function $\Phi$ as
\begin{align}\label{totvardistance}
\Phi\left(\ma{r}_{a},\ma{r}_{b}\right) = \frac{1}{2} \sum_{l=1}^{q_j} \lvert r_{al}- r_{bl} \rvert,
\end{align}
where $r_{al}$ and $r_{bl}$ denote the 
$l$-th element of $\ma{r}_{a}$ and $\ma{r}_{b}$, respectively.

Calculating category dissimilarities using this definition for $\Phi$ is equivalent to the proposal (for categorical variables) by \textcite{AD:2007}. However, as this relationship is not trivial and appears to be unknown, we present this here in some detail.

\subsubsection{Ahmad and Dey's categorical variable distance}
\textcite{AD:2007} argue that the dissimilarity between categories should be computed as a function of their distribution in the overall data set and in co-occurrence with other categories, rather than in isolation. The idea is to take into account co-occurrences of categories when constructing distances. The way they do this, is by considering all combinations of categories of one variable, and selecting the partitioning (that is, a combination of categories) for which the sum of proportions in the two complementary partitions for the two categories is maximal. Following \textcite{AD:2007}, we can define the dissimilarity between categories $a$ and $b$ of variable $i$, with respect to the distribution over the categories of variable $j$, as 
\begin{equation}\label{deltaAD}
\delta(a,b) = \max_{\omega_j} \left({P(\omega_j|a)+P(\bar{\omega_j}|b)-1}\right),
\end{equation}	
where $\omega_j$ and its complement $\bar{\omega_j}$ define a binary partition with respect to the categories of variable $j$, and $P(\omega_j|a)$ denotes the proportion of observations with the category $a$ of variable $i$, corresponding to the set of categories of variable $j$ as defined by $\omega_j$. Note that the term $-1$ is only introduced to fix the upper limit of the dissimilarities at $1$. 
The number of binary partitions for variable $j$, excluding the partitions containing all or no categories, equals $2^{q_j}-2$, where $q_j$ gives number of categories of variable $j$, and hence this number grows exponentially when the number of categories for a variable increases. \textcite{AD:2007} propose an algorithm to calculate their distances. The order of their algorithm is $O(Q^{*2}n + Q^{*2}\bar{q}^3)$, where $Q^{*}$ gives the total number of categories, $n$ is the number of observations and $\bar{q}$ denotes the average number of categories per variable. However, as we show below, and in more detail in Appendix \ref{appen}, for distances between categorical variables, the distance of \textcite{AD:2007} is equivalent to the total variation distance, and calculations using Equation \ref{totvardistance} are much more efficient.

\subsubsection{Equivalence of Ahmad and Dey's distance and the total variation distance between profiles} \label{Sect:ADEquiv} When going from $\delta(a,b)$ to $\delta(b,a)$, the optimal partition $\omega_j$, that is, the combination of categories that maximizes the sum, is simply flipped (i.e., the complement is taken), hence Ahmad and Dey's distance is symmetric: 
\begin{equation*}
\delta(a,b)= \max_{\omega} \left({P(\omega|a)+P(\bar{\omega}|b)-1}\right) = \max_{\omega} \left({P(\omega|b)+P(\bar{\omega}|a)-1}\right) = \delta(b,a),
\end{equation*}
where, for convenience, we dropped the subscripts $j$ for the $\omega$'s. 

As $P(\bar{\omega}|b)=1-P(\omega|b)$ and $P(\bar{\omega}|a)=1- P(\omega|a)$, we have
\begin{align}\label{delta(a,b)_abs}
\delta(a,b)= \max_{\omega} \left({P(\omega|a)-P(\omega|b)}\right)\nonumber\\
= \max_{\omega} \left({P(\omega|b)-P(\omega|a)}\right) \nonumber\\ 
= \max_{\omega} \left\vert {P(\omega|a)-P(\omega|b)} \right\vert. 
\end{align}
Equation (\ref{delta(a,b)_abs}) shows that Ahmad and Dey's distance is equal to finding the maximum difference between all combinations of observed proportions. This implies that we can express this distance as the supremum norm of a vector of differences between probabilities. The total variation distance as defined in (\ref{totvardistance}) can also be defined as the largest difference between probabilities from two probability distributions that can be assigned to the same event. Therefore, the \textcite{AD:2007} distance for categorical variables is equivalent to the total variation distance. For the sake of completeness, we provide a complete proof of the equivalence in Appendix \ref{appen}.



\subsection{Kullback-Leibler divergence between profiles} \label{Sec:KL}
Kullback-Leibler divergence \parencite[KL,][]{KL:1951,K:1959} is an entropy-based measure of dissimilarity between probability distributions. \textcite{LH:2005} define category dissimilarities for the categories of variable $i$ by taking the sum of KL-divergences between the (empirical) conditional probability distributions over all other variables. Using similar notation as before, we can implement this divergence by setting all weights $w_{ij}$ equal to one and by defining $\Phi$ as 
\begin{align*}
\Phi\left( \ma{r}_{a},\ma{r}_{b} \right)  = \sum_{l=1}^{q_j} \left[ r_{al}\log\left(\frac{r_{al}}{r_{bl}} \right) + r_{bl}\log\left( \frac{r_{bl}}{r_{al}} \right) \right],
\end{align*}
where $\log\left( \right)$ is the binary logarithm and $r_{al}$ and $r_{bl}$ denote, as before, $l$-th element of $\ma{r}_{a}$ and $\ma{r}_{b}$, respectively. It is important to note that $KL$ is not symmetric. Hence, distance calculations using $\bm{\Delta}_{KL}$ may result in non-symmetric distances. 

\subsection{$\chi^{2}$-distance between profiles} \label{Sectchi}
A distance for categorical data, that has a strong link to the data visualization technique correspondence analysis, is the chi-squared distance. There exist several forms and implementations of the chi-square distance that differ with respect to the chosen standardization. That, is, chi-squared distance considers the squared differences between proportions divided by the expected proportions. For an $n \times p$ contingency matrix $\ma{F}_j= \ma{Z}_{i}^{\prime}\ma{Z}_{j}$, the squared $\chi^2$-distance between rows $a$ and $b$ can be defined as 
\begin{align}\label{Chi2}
s \sum_{l=1}^{p} {\frac{1}{f_{\bullet l}} \left(\frac{f_{al}}{f_{a\bullet}}- \frac{f_{bl}}{f_{b\bullet}} \right)^2 ,}
\end{align}
where $s$ is the sum of all elements of $\ma{F}$ and $\bullet$ denotes the summation in the appropriate dimension (rows or columns) of the matrix \parencite[see, e.g.,][p.266]{G:1990}. In our notation, we can implement the chi-squared distances as category dissimilarities by defining
\begin{align*}
\Phi\left(\ma{r}_{a},\ma{r}_{b}\right) = \sum_{l=1}^{q_j}{ \frac{1}{p_{l}}\left(r_{al} - r_{bl}\right)^{2}},
\end{align*}
where we dropped the constant $s$, $p_l$ corresponds to the $l$-th element of the $j$-th block of $\ma{P}_d$ and, as before, $r_{al}$ and $r_{bl}$ denote the
$l$-th element of $\ma{r}_{a}$ and $\ma{r}_{b}$, respectively.

\section{Supervised association-based distances} \label{SectSupervised}
In a supervised setting, where we want to assign observations to classes (i.e., categories) for one variable, say $y$, based on observations on categorical variables $x_{j}$ where $j=1, \dots, Q$, we can define a supervised variant of association-based categorical variable distances. That is, we can define category dissimilarities that take into account the association between variables $y$ and $x$. Next, we can make predictions using either the $K$-nearest neighbors or a distance-based clustering method, where we fix the number of clusters to the number of classes and do a post-hoc comparison of clusters and classes. That is, we match the clusters to the true classes and assign labels accordingly. 

To define supervised association-based distances we create an $n \times c$ indicator matrix $\ma{Z}_{y}$, where $c$ corresponds to the number of classes of $y$. If we add, to the right, this indicator matrix to $\ma{Z}$ and insert this supplemented $\ma{Z}$ into Equations (\ref{P}) through (\ref{R}), we can calculate category dissimilarities using Equations (\ref{Phi_ij}) and (\ref{genformula}). Note that in this new setting, we have $Q+1$ variables and consequently $Q+1$ association-based category dissimilarity matrices $\bm{\Delta}^{i}$. However, the $(Q+1)$-th diagonal block gives the category dissimilarities between the categories of the $y$ variable. In a supervised setting, a category (class) of $y$ is to be predicted based on data from the other $Q$ variables. The category dissimilarities for $y$ should therefore not be used. This is easily achieved by simply ignoring these in the overall block diagonal category dissimilarity matrix $\bm{\Delta}$. That is, we construct $\bm{\Delta}$ by collecting only the first $Q$ category dissimilarity matrices on its diagonal.

As before, our framework allows for great flexibility in how to incorporate the information of variable $y$. In particular, the dissimilarity functions $\Phi^{ij}$ and the weights $w_{ij}$ are pair-specific. One could, as suggested in Section \ref{SectAssociationBasedDist}, set all weights $w_{ij}$ equal to $1$, so that the category dissimilarities take into account all associations. We refer to this choice as \say{full supervised} dissimilarity. 

Alternatively, in a supervised setting, one may choose to have the category dissimilarities depend only on the association with the variable $y$. This corresponds to the choice $w_{ij} = 1$ for $j=Q+1$ and $0$ for all other pairs. In this case, category dissimilarities may better discriminate with respect to the classes of $y$. We refer to this choice as \say{supervised} dissimilarity. Note that both supervised variants require a choice of association-based dissimilarity functions $\Phi^{ij}$, for all pairs of variables.

\section{Aggregation and dissimilarity tuning} \label{Sect:Tuning}
Our general framework introduced in Section \ref{SectCatDistances} allows for great flexibility in the implementation of distances between categorical variables. In particular, in the previous sections, we introduced a selection of category dissimilarity measures. However, there are many more options. For example, all $46$ functions available in the R package \texttt{philentropy}, described in \textcite{D:2018}, can be used for association-based functions. Moreover, as is clear from Definition (\ref{genformula}, all separate category dissimilarity measures can be combined and aggregated according to the researcher's preferences. How to exactly determine which category dissimilarity and aggregation strategy is the most appropriate is non-trivial, and this choice may depend on the properties of the data and the research objectives. 

In several distance-based methods, for example cluster analysis and multidimensional scaling, a clear measure of fit is not available, as the methods tend to be primarily exploratory. That is, the goal is to find and interpret patterns in the data. As the interpretability of a solution is not easily quantified, validation is typically not trivial. If, however, a measure of fit can be calculated, we can use this to select an aggregation and category-dissimilarity definition. That is, we can apply several aggregation and category dissimilarity definitions, and compare the fit for each of them by considering the selected measure.

In a supervised classification setting, where we have a data set for which the true classes are known, we can assess the fit by comparing true classes with \say{predicted} classes. A choice for aggregation and category dissimilarity definitions, can then be made based on the discrepancy between these. Therefore, the aggregation state (that is, the weights $w_{ij}$) and the category dissimilarity function (that is, $\Phi^{ij}$) can be treated as tuning parameters. Note, however, that Equation \ref{genformula} allows many combinations and some choices need to be made to restrict the total search space.


\section{Applications} \label{SectApplications}
To illustrate how our general framework can be used in practice, we consider distance-based methods for supervised and unsupervised learning. In a supervised setting, a distance-based approach is $K$-nearest neighbors averaging, which can be used in regression and classification problems: each new observation is labeled according to a set of $K$ {\em close} training points (neighbors). In an unsupervised setting, distance-based cluster analysis aims to assign observations to groups (clusters) for which the within-cluster distances are small, whereas the distances between clusters are large. 

As a general setup, we consider nine different labeled data sets (see Table \ref{Tab:1}), all available via the UCI Machine Learning repository\footnote{\url{https://archive.ics.uci.edu/ml/index.php}}. Each data set is split into five folds, for cross-validation. On the training folds, a block diagonal matrix $\bm{\Delta}$ of pair-wise category dissimilarities is calculated for each of the reviewed dissimilarity measures (see Table \ref{Tab:2}). The test fold is used for performance assessment of the considered methods. The performance metric depends on the considered method:

\begin{itemize}
\item {\em Accuracy} of the nearest neighbors classifier: proportion of the test observations correctly classified \parencite{Me:1978};
\item {\em Adjusted Rand Index} \parencite[{\em ARI},][]{HA:1985} comparing the cluster allocation of the test observations to the {\em true} cluster allocation (the labels).
\end{itemize}

The procedure is iterated until each fold is used as test. The whole cross-validation process is repeated $10$ times, for different random splits. 
\begin{table}[ht]
\centering
\caption{Data set information} \label{Tab:1}
\begin{tabular}{lrrr}
\toprule
Dataset & $n$   & $p$   & \# clusters \\ 
\hline
australian      & 690       & 8     & 2 \\
balance         & 625       & 4     & 3 \\
cars            & 1728      & 6     & 4 \\
lympho          & 148       & 18    & 4 \\
soybean (large) & 307       & 35    & 19 \\
tae             & 151       & 5     & 3 \\
tictac          & 958       & 9     & 2 \\
vote            & 435       & 16    & 2 \\
wbcd            & 699       & 9     & 2 \\  
\bottomrule    
\end{tabular}
\end{table}

\begin{table}[ht]
\centering
\caption{Category dissimilarity measures considered in the experiments} \label{Tab:2}
\begin{tabular}{ll}
\toprule
Independent & Association-based \\ 
\hline
SM (Sec. \ref{Sect:SM}) &  TVD (Sec. \ref{Sect:TVD})\\
Eskin (Sec. \ref{Sec:Eskin}) & KL (Sec. \ref{Sec:KL}) \\
Lin (Sec. \ref{Sec:Lin}) & KL (Sec. \ref{Sec:KL})\\
IOF (Sec. \ref{Sec:IOF}) &  Supervised TVD, Supervised TVD-full (Sec. \ref{SectSupervised})\\
OF (Sec. \ref{Sec:OF}) & \\
Goodall 3 and 4 (Sec. \ref{Sect:Goodall}) & \\
VE, VM (Sec. \ref{Sec:VEVM})\\
\bottomrule
\end{tabular}
\end{table}

\subsection{$K$-nearest neighbors of categorical data}

The KNN classification of the test observations is based on the calculation of the distance between each test observation and the training observations, as described in Section \ref{dist_sets}. Let $\bm{\Delta}_{train}$ denote the category dissimilarity matrix, where the subscript $train$ indicates that if the category dissimilarities are data dependent, only observations of the training set were used. Furthermore, $\ma{Z}_{test}$ and $\ma{Z}_{train}$ are the indicator matrices of the test and training observations, respectively. The distances of interest are in the columns of
\begin{equation*}
\ma{Z}^{}_{train}\bm{\Delta}_{train}\ma{Z}_{test}^{\prime}
\end{equation*}
and the nearest neighbors for the $j$-th test observation are the $K$ smallest values in the $j$-th column.

Since the lower the number of considered neighbors, the higher the flexibility of the classifier, $K$ is a hyper-parameter. We tune this hyper-parameter using the repeated cross-validation validation procedure described in Section \ref{SectApplications}. In particular, we consider values for $K \in \left\{1,3,5,9,15,21\right\}$ and, for each data set/category dissimilarity combination, the value of $K$ is chosen that minimizes the cross-validation estimate of the classifier's test accuracy.

\begin{figure}[ht]
\center
\includegraphics[scale=.62]{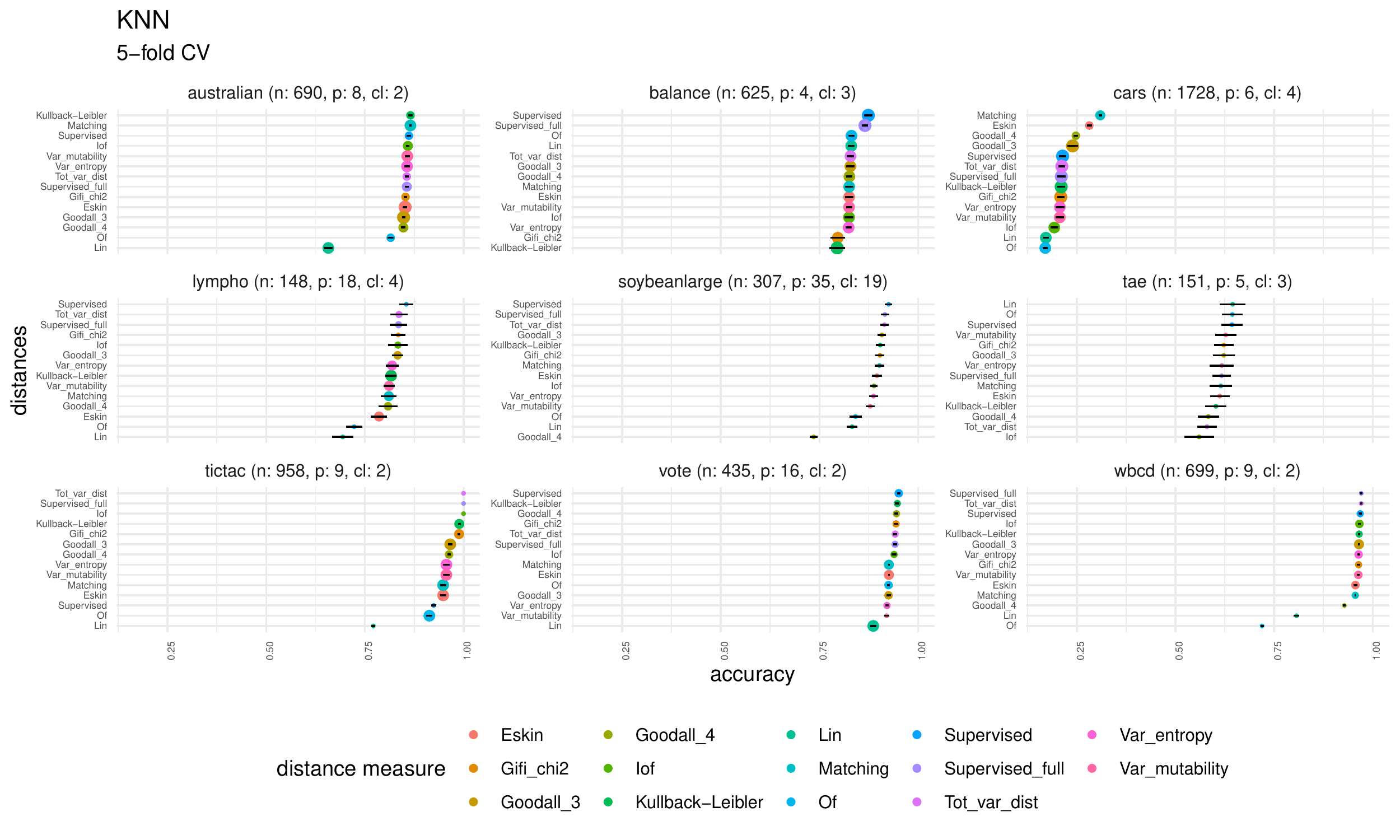}
\caption{KNN classification accuracy for each considered distance measure and data set}
\label{fig:knn}
\end{figure}

Figure \ref{fig:knn} presents the accuracy assessment of the tuned KNN classifier for each considered data set and for each considered category dissimilarity definition. In each panel, the position of each point corresponds to the accuracy obtained using the indicated category dissimilarity definition; the size of each point is proportional to the tuned value of the hyper-parameter $K$. The lines centered at each point span twice the standard deviation of the accuracy over the 10 replicates. For each data set, the distances are reported in descending order, highlighting the best performing ones. For some data sets, e.g., $australian$, $vote$ and $wbcd$, the accuracy is high almost irrespective to the chosen distance, with no variability over the 10 cross-validation replicates. For smaller data sets, such as $lympho$ and $tae$, there is more variability over the 10 cross-validation replicates, as expected. 

\subsection{Partitioning around medoids}
Partitioning around medoids (PAM) is an iterative clustering procedure that takes as input a matrix of pair-wise distances between a set of observations. Within a cluster, a medoid corresponds to the {\em median} observation, just like a centroid in K-means corresponds to the mean. The starting set of the $K$ medoids is random, and each observation is assigned to the closest medoid; given the allocation of the obtained clusters, the medoids are updated accordingly. The procedure stops iterating when there are no changes in the set of medoids. Although we are working in an unsupervised context, the performance of PAM on each data set/category dissimilarity combination can be assessed via cross-validation by calculating distances based on the supervised association dissimilarities; this also allows for consistency with the KNN-based application.

In particular, for each data set and each category dissimilarity definition, a medoids set is obtained by applying PAM to the training data. That is,
\begin{equation*}
\ma{D} = \ma{Z}^{}_{train}\bm{\Delta}_{train}\ma{Z}^{\prime}_{train}    
\end{equation*}
where for ${\bm \Delta_{train}}$ we consider all category dissimilarity matrices described in Sections \ref{Sect:IndependentDistances} and \ref{SectAssociationBasedDist} and reported in Table \ref{Tab:2}. Next, we compute the test observation-to-medoid distance matrix as follows,
\begin{equation*}
\ma{Z}^{}_{medoid}\bm{\Delta}_{train}\ma{Z}_{test}^{\prime},
\end{equation*}
and assign each test observation to the cluster corresponding to the nearest medoid.

\begin{figure}[ht]
\centering
\includegraphics[scale=.62]{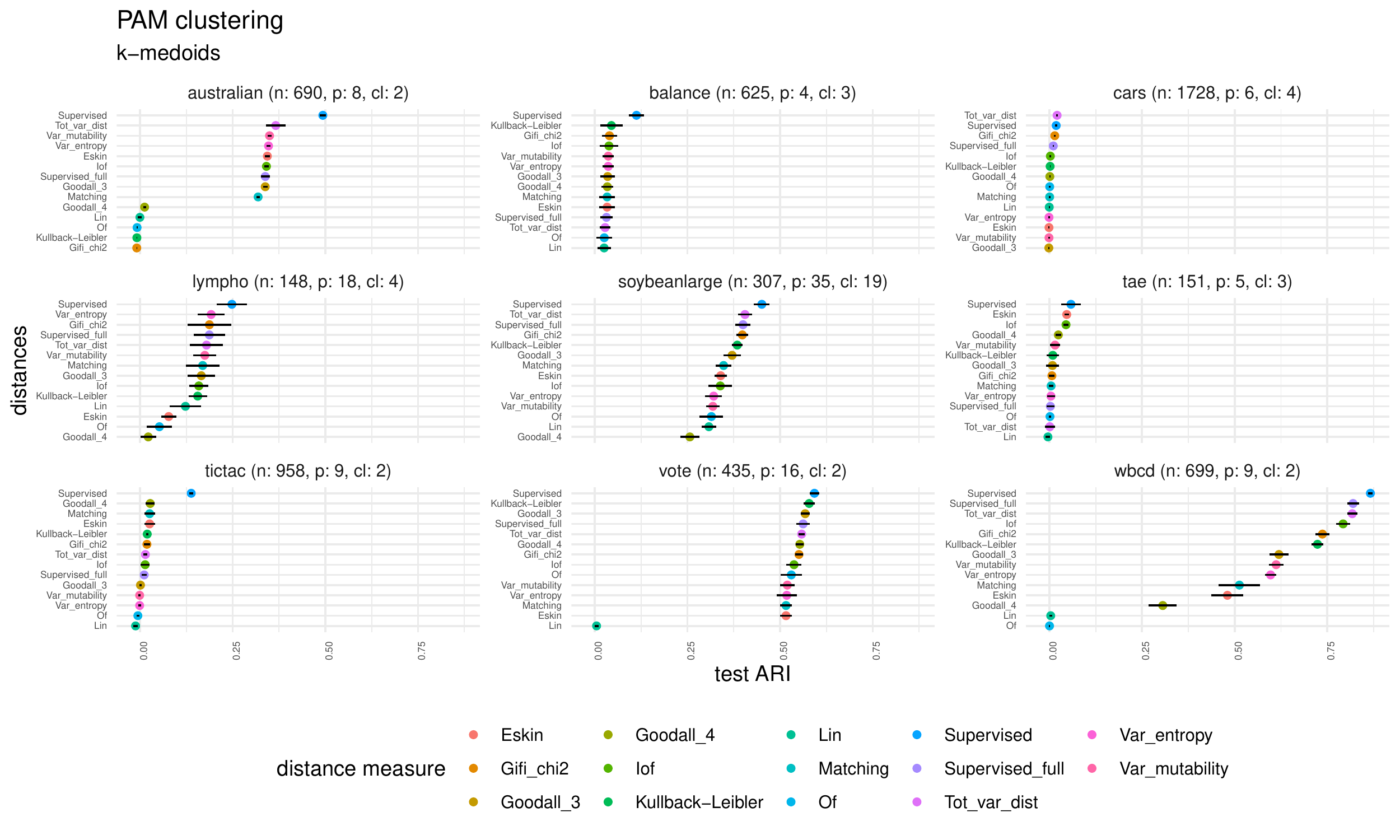}
\caption{ARI results on K-medoids}
\label{fig:pam}
\end{figure}

The results are reported in Figure \ref{fig:pam}. We observe that, with the exception of the {\em cars} data set, for which performance of all methods is poor, the supervised association-based distance generally performs well.

In general, the KNN and PAM results lead to the following conclusions. 
\begin{itemize}
\item Data sets for which classification accuracy is high in the supervised setting also have higher ARI values in the unsupervised setting.
\item The choice of category dissimilarity does not appear to impact classification accuracy when the overall performance of the method is very poor (for example, for {\em cars}) or very good (e.g., for {\em wbcd}).
\item In an unsupervised setting, association-based measures seem to provide an edge: in {\em wbcd}, five out of the six measures with an ARI value above $0.75$ are association-based.
\end{itemize}

Note that extended results and the R code to reproduce them are available online \footnote{\url{https://alfonsoiodicede.github.io/blogposts_archive/distances_experiment_superv_unsuperv.html}}. 

\section{Conclusion} \label{SectConclusion}
In this paper, we propose a general framework for implementing distances between categorical variables in a flexible, efficient and transparent manner. In detail, we show that both independent and association-based distances can be incorporated in our framework. The latter can therefore be used to implement several existing measures, as well as to easily introduce new highly customizable ones. 

Our proposal is valuable from a theoretical perspective because it allows assessing the differences between dissimilarity measures by simplifying the wide variety of notation and definitions used in the literature, and therefore making their implementation much more transparent. From an applied perspective, since the proposed framework is not method- or application-specific, it can allow the definition of problem-specific dissimilarity measures. With respect to this, it is important to outline that in a supervised context our proposal can be used to define measures that consider associations with a response variable. 

For the independent category dissimilarities, the definitions described in Section \ref{Sect:IndependentDistances} (with the exception of the ordered category dissimilarities) are implemented in the \texttt{nomclust} R package \parencite{SR:2015}. However, association-based measures are not implemented in this 
package. In the
\texttt{catdist} package\footnote{Available on GitHub at \url{https://github.com/alfonsoIodiceDE/catdist_package} and will soon be released on CRAN.}  the independent as well as the association-based measures presented in this paper 
are implemented.

To illustrate the importance of selecting the \say{best} or the \say{most appropriate} distance for the problem at hand, we used our framework in both supervised (via KNN) and unsupervised (via PAM) contexts. Applications on real-world data sets revealed that choosing a specific measure will not affect neither classification accuracy nor clustering performance, respectively, when the variables have no discriminatory power,  there is no strong cluster structure in the data, or KNN/ PAM are not appropriate methods for the problem at hand. Similarly, there are cases where all measures perform equally well. Putting extreme scenarios aside, choosing the \say{most appropriate} measure can lead to a classification / clustering improvement and association-based measures outperformed, in many cases, independent category measures. A further, more structured study, using synthetic data as well as a larger collection of empirical data sets, is needed to appraise this claim. Such a study is beyond the scope of this paper.

By using the general framework proposed in this paper, new or customized distances can easily be implemented. In fact, the supervised total variation distances introduced in Section \ref{SectSupervised}, for example, are \say{new} measures. However, rather than introducing and appraising new measures, it might be more interesting to consider a more systematic comparison of the strengths and weaknesses of different dissimilarity measures for categorical variables that are already available in the literature. 


\begin{appendices} 
\section{} \label{appen}
Here we prove that the definition of category dissimilarities using total variance, as in Equation (\ref{totvardistance}), is equivalent to the definition of category dissimilarities proposed in \textcite{AD:2007}. Recall that, as explained in Section \ref{Sect:ADEquiv}, \textcite{AD:2007} define the dissimilarity between categories $a$ and $b$ of variable $i$, with respect to the distribution over the categories of variable $j$, as 
\begin{equation} \label{AppendixAD1}
\delta(a,b) = \max_{\omega_j} \left({P(\omega_j|a)+P(\bar{\omega_j}|b)-1}\right),
\end{equation}	
where $\omega_j$ and its complement $\bar{\omega_j}$ define a binary partition with respect to the categories of variable $j$, and $P(\omega_j|a)$ denotes the proportion of observations with the category $a$ of variable $i$, corresponding to the set of categories of variable $j$ as defined by $\omega_j$. In Section \ref{Sect:ADEquiv}, we showed that Equation (\ref{AppendixAD1}) is equivalent to
\begin{align} \label{AppendixAD2}
\delta(a,b)= \max_{\omega_j} \left\vert {P(\omega_j|a)-P(\omega_j|b)} \right\vert.
\end{align}
Let $\ma{K}_{q_{j}}$ be a design matrix that defines all, except the empty and complete, binary partitions for the $q_{j}$ categories of variable $j$. Hence, $\ma{K}_{q_{j}}$ is a $q_j \times q_{j}^{\star}$ matrix of zeros and ones, where $q_{j}^{\star} = \sum_{l=1}^{q_{j}-1}{\binom{q_j}{l}}=2^{q_j}-2$. We can re-express Equation (\ref{AppendixAD2}) as
\begin{equation*}
\delta(a,b) = \left\lVert \ma{K}^{\prime}_{q_j} \left(\ma{r}_{a} - \ma{r}_{b}\right) \right\rVert_{\infty} =  \left\lVert \ma{K}^{\prime}_{q_j} \ma{d}_{ab} \right\rVert_{\infty},
\end{equation*}
where $\ma{d}_{ab}=\left(\ma{r}_{a} - \ma{r}_{b}\right)$ and $\left\lVert \ma{x} \right\rVert_{\infty} $ denotes the supremum norm of vector $\ma{x}$, that is, the maximum element of $\ma{x}$ in absolute value.

The number of columns of $\ma{K}_{q_j}$ and hence the size of the vector from which we need to take the norm, grows exponentially with the number of categories. For example, for $q_j=4$, $q^{\star}_j=14 $ but for $q_j=8$ we have $q^{\star}_j=254$. When considering binary partitions, only half of these combinations are needed. Still, when the number of categories of the categorical variables is not too small, considering all combinations becomes computationally expensive. 

A more efficient way to calculate the distances between categories $a$ and ${b}$ can be obtained using the following relationship
\begin{equation}\label{equivalence}
\left\lVert \ma{K}^{\prime}_{q_j} \ma{d}_{ab} \right\rVert_{\infty} = \frac{1}{2}\left\lVert \ma{d}_{ab} \right\rVert_{1},
\end{equation}
where $\left\lVert \ma{x} \right\rVert_{1} $ denotes the $L_{1}$ norm of vector $\ma{x}$, that is 
\begin{equation*}
\left\lVert \ma{x} \right\rVert_{1} = \sum_i \left\lvert x_i \right\rvert.
\end{equation*}

To see that Equation (\ref{equivalence}) holds, note that the maximum, in absolute value, for the combinations of elements of $\ma{d}_{ab}$ is obtained by selecting the combination consisting of elements that have the same sign. Furthermore, as the sum of elements of $\ma{d}_{ab}$ equals zero, that is:

\begin{equation*}
\ma{d}_{ab}\ma{1}_{q_{j}} = \left(\ma{r}_{a} - \ma{r}_{b}\right)\ma{1}_{q_{j}} = 0,
\end{equation*}
it immediately follows that the sum of all positive elements equals the sum of all negative values. Therefore,
\begin{equation*}
\left\lVert { \ma{K}^{\prime}_{q_{j}} \ma{d}_{ab}} \right\rVert_{\infty} = \sum_{l: d_{l}>0} {\left\lvert d_{l} \right\rvert} = \sum_{l: d_{l}<0} {\left\lvert d_{l} \right\rvert}, 
\end{equation*}
where $d_{l}$ denotes the $l$-th element of $\ma{d}_{ab}$. Finally, as 
\begin{equation*}
\left\lVert {\ma{d}_{ab}} \right\rVert_{1} = \sum_l {\left\lvert d_{l} \right\rvert} = \sum_{l: d_{l}>0} {\left\lvert d_{l} \right\rvert} + \sum_{l: d_l<0} {\left\lvert d_l \right\rvert}
\end{equation*}
the equivalence in Equation (\ref{equivalence}) immediately follows, and we can express Ahmad and Dey's distance between categories $a$ and $b$ with respect to the categories of variable $j$, as
\begin{equation}\label{delta(a,b)_L1}
\delta(a,b)= \frac{1}{2}\left\lVert \ma{d}_{ab} \right\rVert_{1} = \frac{1}{2}\left\lVert \ma{d}_{ab} \right\rVert_{1} = \frac{1}{2} \sum_{l=1}^{q_j} \lvert r_{al} - r_{bl} \rvert .
\end{equation}
Comparing Equations \ref{totvardistance} and \ref{delta(a,b)_L1} shows that Ahmad and Dey's distance is equivalent to the total variation distance introduced in Section \ref{Sect:TVD}.
\end{appendices}

\section*{Funding}
The authors received no financial support for the research, authorship, and/or publication of this article. 

\section*{Conflict of interest}
The authors declare that they have no conflict of interest.

\section*{Data availability}
Extended results and the code to reproduce them are available online at \url{https://alfonsoiodicede.github.io/blogposts_archive/distances_experiment_superv_unsuperv.html}. The data used in this study were downloaded from the UCI repository \parencite{DG:2017} and are also available in the \texttt{catdist} package available on GitHub at \url{https://github.com/alfonsoIodiceDE/catdist_package}.

\printbibliography

\end{document}